\documentclass[conference]{IEEEtran}
\IEEEoverridecommandlockouts
\usepackage{cite}
\usepackage{amsmath,amssymb,amsfonts}
\usepackage{algorithmic}
\usepackage{graphicx}
\usepackage{textcomp}
\usepackage{xcolor}
\def\BibTeX{{\rm B\kern-.05em{\sc i\kern-.025em b}\kern-.08em
    T\kern-.1667em\lower.7ex\hbox{E}\kern-.125emX}}

\usepackage{tikz}
\usepackage{bm}
\usepackage{caption}
\usepackage{subcaption}

\usepackage{amsmath,amsfonts,amssymb,amsthm}
\usepackage{mathtools}
\usepackage{commath}
\usepackage{bbm}
\usepackage{hyperref}
\usepackage{float}

\newcommand{\csim}{\text{sim}}

\begin{document}

\title{A Contrastive Learning Approach to Auroral Identification and Classification \\
\thanks{This work is supported by the National Science Foundation grant OIA--1920965.}
}

\author{\IEEEauthorblockN{1\textsuperscript{st} Jeremiah W. Johnson}
        \IEEEauthorblockA{\textit{Dept. of Applied Engineering \& Sciences} \\
\textit{University of New Hampshire}\\
Manchester, NH, USA \\
\texttt{jeremiah.johnson@unh.edu}}
\and
\IEEEauthorblockN{2\textsuperscript{nd} Swathi Hari}
\IEEEauthorblockA{\textit{Dept. of Applied Engineering \& Sciences} \\
\textit{University of New Hampshire}\\
Manchester, NH, USA \\
\texttt{spj322@wildcats.unh.edu}}
\and
\IEEEauthorblockN{3\textsuperscript{rd} Donald Hampton}
\IEEEauthorblockA{\textit{Geophysical Institute} \\
\textit{University of Alaska--Fairbanks}\\
Fairbanks, AK, USA \\
\texttt{dhampton@alaska.edu}}
\and
\IEEEauthorblockN{4\textsuperscript{th} Hyunju Connor}
\IEEEauthorblockA{\textit{Geophysical Institute} \\
\textit{University of Alaska--Fairbanks}\\
Fairbanks, AK, USA \\
\texttt{hkconnor@alaska.edu}}
\and
\IEEEauthorblockN{5\textsuperscript{th} Amy Keesee} 
\IEEEauthorblockA{\textit{Department of Physics} \\
\textit{University of New Hampshire}\\
Durham, NH, USA \\
\texttt{amy.keesee@unh.edu}}
}

\maketitle

\begin{abstract}
        Unsupervised learning algorithms are beginning to achieve accuracies comparable to their supervised counterparts on benchmark computer vision tasks, but their utility for practical applications has not yet been demonstrated. In this work, we present a novel application of unsupervised learning to the task of auroral image classification. Specifically, we modify and adapt the Simple framework for Contrastive Learning of Representations (SimCLR) algorithm to learn representations of
        auroral images in a recently released auroral image dataset constructed using image data from Time History of Events and Macroscale Interactions during Substorms (THEMIS) all--sky imagers. We demonstrate that (a) simple linear classifiers fit to the learned representations of the images achieve state--of--the--art classification performance, improving the classification accuracy by almost 10 percentage points over the current benchmark; and (b) the learned representations naturally cluster into more clusters than exist manually assigned categories, suggesting that existing
        categorizations are overly coarse and may obscure important connections between auroral types, near--earth solar wind conditions, and geomagnetic disturbances at the earth's surface. Moreover, our model is much lighter than the previous benchmark on this dataset, requiring in the area of fewer than 25\% of the number of parameters. Our approach exceeds an established threshold for operational purposes, demonstrating readiness for deployment and utilization.
\end{abstract}

\begin{IEEEkeywords}
        Unsupervised learning, representation learning, heliophysics, aurora.
\end{IEEEkeywords}

\section{Introduction}\label{section:introduction}

Understanding the interaction of the solar wind with the Earth’s magnetosphere and upper atmosphere is one of the key topics in heliophysics~\cite{2014scienceplan,2013nrc}. Aurora Borealis and Aurora Australis are formed when high--energy electrons and ions precipitate from the magnetosphere to the upper atmosphere due to the coupling of solar wind and magnetosphere. These auroral features at a scale of 10 -- 10,000km hold the key information of the magnetospheric coupling processes at a scale of 1 -- 1,000 earth radii.  Since it is difficult to cover the entire magnetosphere with a limited number of satellites, ground--based auroral images have been critical to understanding the propagation of the solar wind energy to near--Earth space~\cite{akasofu1964development,akasofu1981energy,clausen2014thermospheric}. 

With decades of ground and space observations, global aurora patterns of 1,000 -- 10,000km and their general mechanisms are well understood~\cite{newell2009diffuse}. However, the morphology of smaller--scale aurora forms (less than 1000km) and their connections to mid-- to large--scale magnetospheric dynamics (less than 100 earth radii) are still under question, due both to the complexity of aurora features and the abundance of aurora images collected. A better understanding of local--scale auroral morphology is critical to improving our understanding of the solar wind--magnetosphere--upper atmosphere interaction.  

The heliophysics community has over decades amassed vast collections of images of Aurora Borealis and Aurora Australis. To date, the vast majority of this data is unclassified. Classifying this data is a first and important key step to identify local--scale auroral features and thus to enable a deeper analysis on the link between auroral features and magnetospheric dynamics. 

%
%

\section{Related Work}\label{section:related}
Automatic classification of auroral images has been studied in the past; tradition computer vision approaches reliant on the construction of hand--designed features include~\cite{syrjasuo2002analysis,syrjasuo2004diurnal,syrjasuo2007automatic,rao2014automatic}. While most of these approaches are only effective when limited to binary \emph{aurora/no aurora} classification, the approach of~\cite{yang2012auroral} using a hidden Markov model and incorporating temporal dynamics is able to identify four distinct categories of aurora with a positive detection rate of up to 85\%. 

\begin{figure*}[!t]
        \centering
        \begin{subfigure}[b]{0.15\textwidth}
                \centering
                \includegraphics[width=\textwidth]{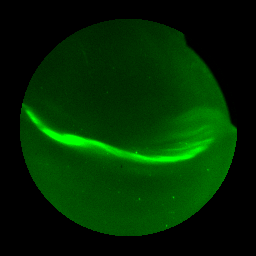}
                \label{fig:clausen-arc-top}
        \end{subfigure}
        \hfill
        \begin{subfigure}[b]{0.15\textwidth}
                \centering
                \includegraphics[width=\textwidth]{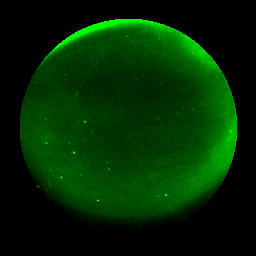}
                \label{fig:clausen-diffuse-top}
        \end{subfigure}
        \hfill
        \begin{subfigure}[b]{0.15\textwidth}
                \centering
                \includegraphics[width=\textwidth]{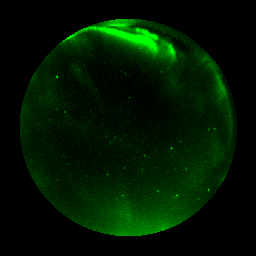}
               \label{fig:clausen-discrete-top}
        \end{subfigure}
        \hfill
        \begin{subfigure}[b]{0.15\textwidth}
                \centering
                \includegraphics[width=\textwidth]{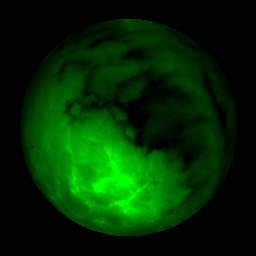}
                \label{fig:clausen-cloudy-top}
        \end{subfigure}
        \hfill
        \begin{subfigure}[b]{0.15\textwidth}
                \centering
                \includegraphics[width=\textwidth]{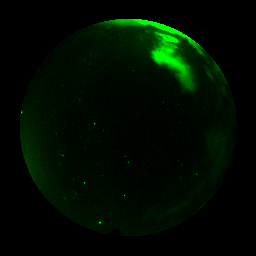}
                \label{fig:clausen-moon-top}
        \end{subfigure}
        \hfill
        \begin{subfigure}[b]{0.15\textwidth}
                \centering
                \includegraphics[width=\textwidth]{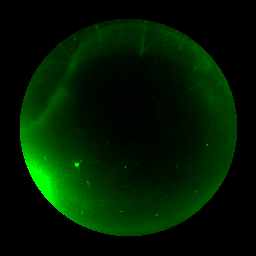}
                \label{fig:clausen-clear-top}
        \end{subfigure}

         \begin{subfigure}[b]{0.15\textwidth}
                \centering
                \includegraphics[width=\textwidth]{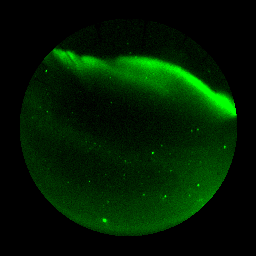}
                \caption{Arc}
                \label{fig:clausen-arc}
        \end{subfigure}
        \hfill
         \begin{subfigure}[b]{0.15\textwidth}
                \centering
                \includegraphics[width=\textwidth]{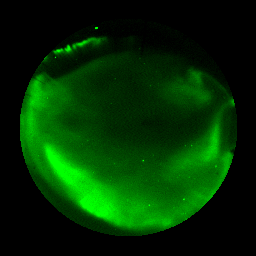}
                \caption{Diffuse}
                \label{fig:clausen-diffuse}
        \end{subfigure}
        \hfill
         \begin{subfigure}[b]{0.15\textwidth}
                \centering
                \includegraphics[width=\textwidth]{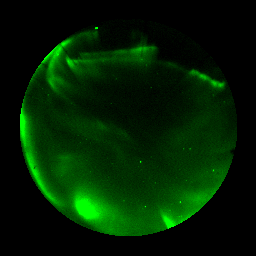}
                \caption{Discrete}
                \label{fig:clausen-discrete}
        \end{subfigure}
        \hfill
         \begin{subfigure}[b]{0.15\textwidth}
                \centering
                \includegraphics[width=\textwidth]{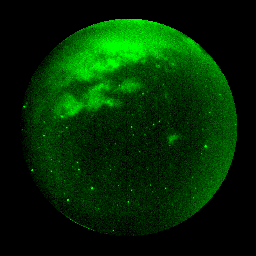}
                \caption{Cloudy}
                \label{fig:clausen-cloudy}
        \end{subfigure}
        \hfill
         \begin{subfigure}[b]{0.15\textwidth}
                \centering
                \includegraphics[width=\textwidth]{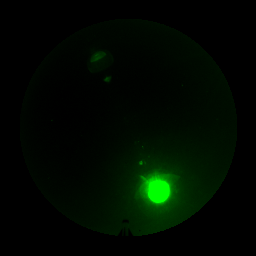}
                \caption{Moon}
                \label{fig:clausen-moon}
        \end{subfigure}
        \hfill
         \begin{subfigure}[b]{0.15\textwidth}
                \centering
                \includegraphics[width=\textwidth]{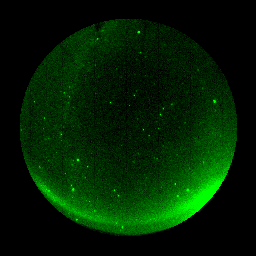}
                \caption{Clear}
                \label{fig:clausen-clear}
        \end{subfigure}

        \caption{Random samples of images from the OATH dataset showing the distinct categories of aurora.}
        \label{fig:image-samples}
\end{figure*}

Computer vision is now dominated by convolutional neural network (CNN)--based algorithms that have achieved remarkable performance on a wide range of challenging tasks. Following the general trend in computer vision, there have been several recent efforts to use CNN--based models for automatic auroral classification. In~\cite{clausen2018automatic}, Clausen \emph{et. al.} manually label 5,824 images of aurora obtained from THEMIS all--sky imagers into the distinct categories \emph{Arc},
\emph{Discrete}, \emph{Diffuse}, \emph{Clear/No Aurora}, \emph{Cloudy}, and \emph{Moon}. Using an Inception model~\cite{szegedy2015going} pretrained
on the ImageNet dataset~\cite{deng2009imagenet,sharif2014cnn}, they extract feature vectors and fit a ridge classifier to them to obtain an 82\% accuracy. In~\cite{kvammen2020auroral}, the authors manually classified 3,846 auroral images into the highly imbalanced categories \emph{Arc aurora}, \emph{Auroral breakup}, \emph{Colored aurora}, \emph{Discrete aurora}, \emph{Edge aurora}, \emph{Faint Aurora}, \emph{Patchy Aurora}, then compare classification results from a
range of machine learning algorithms, including CNN--based architectures.   

There are two closely related challenges inherent to using conventional CNN--based approaches to auroral image analysis. First, supervised learning methods require a substantial quantity of unambiguous, high--quality labeled data be available for training. While there exists massive amounts of high--quality auroral image data, to date very little of this data has been labeled. The use of ImageNet--pretrained CNNs as in~\cite{clausen2018automatic} presents a possible alternative, as pretraining is often effective at reducing the amount of data necessary to train a CNN to convergence~\cite{razavian2014cnn}, but pretraining only partially addresses the challenge of limited data and may introduce other issues and biases~\cite{geirhos2018imagenet}. Secondly and more importantly, manually labeling auroral images itself presents additional challenges beyond the time and expense required: labels for auroral images are often chosen based on subjective, qualitative judgements, and there is frequent disagreement between researchers regarding what an appropriate label set is. Complicating the picture further is the fact that even after a label set is decided on it is common for all--sky auroral images to contain more than one type of aurora, raising the additional question of how to decide what the correct classification should be.

In the past few years, unsupervised machine learning algorithms such as \cite{chen2020simple}, \cite{caron2020unsupervised}, \cite{Dai_2021_CVPR} have begun to achieve levels of performance rivaling those of their supervised counterparts on benchmark tasks and datasets, including CIFAR10, CIFAR100, and ImageNet. However, the utility and efficacy of recent unsupervised models for practical tasks and real--world data is not yet well established. Unsupervised models afford an opportunity to address the challenges just described in auroral imaging simultaneously: the learned representations of auroral images may be useful for quantitatively determine meaningful labels, and can be used to apply such labels automatically, facilitating the creation of much larger labeled auroral image datasets. Our contribution in this work is as follows:

\begin{itemize}
        \item We modify and adapt the \emph{Simple Framework for Contrastive Learning of Representations} (SIMCLR) model introduced in~\cite{chen2020simple} and apply it to a publicly available auroral image dataset released with the publication of~\cite{clausen2018automatic}, consisting of 5,824 images of aurora obtained from THEMIS all--sky imagers~\cite{donovan2006themis} and manually categorized into the distinct categories \emph{Arc}, \emph{Discrete}, \emph{Diffuse}, \emph{Clear/No Aurora}, \emph{Cloudy}, and \emph{Moon}. Our approach leads to sufficiently informative representations to enable a simple linear classifer to obtain state--of--the--art classification performance across these categories, reducing the top--1\% error rate from the previous benchmark by almost 10 percentage points. Moreover, our model obtains this performance while requiring less than 25\% of the number of parameters of the model used to obtain the previous state--of--the--art.
        \item We demonstrate that the representations our model learns for these images cluster naturally into more categories than exist manually assigned ground--truth labels, suggesting that the current labels may be overly coarse and may obscure important information about auroral morphology that may be useful in understanding the connection between auroral morphology and magnetospheric dynamics.
\end{itemize}

The rest of the paper is organized as follows: in Section \ref{section:methodology}, we describe our model, training strategy, and results. In Section \ref{section:conclusion}, we conclude with a discussion of future work.

\section{Methodology}\label{section:methodology}

\subsection{Data}\label{section:data}

The dataset used in this study consists of 5,824 images collected from various THEMIS all--sky imagers. This dataset was initially constructed for the study presented in~\cite{clausen2018automatic}, and is publicly available at \url{http://tid.uio.no/plasma/oath/oath_v1.1_20181026.tgz}. We refer to this dataset as the Oslo Auroral THEMIS (OATH) dataset. 

Each image $\bm{x}$ in the dataset was cropped by 15\% to remove pixels corresponding to very low elevation angles and then scaled to enhance dim features by applying the following formula:
\begin{equation}
        x_{ij} = \max_{i, j}(\min_{i, j}(\frac{x_{ij} - m_{\bm{x}}}{M_{\bm{\tilde{x}}}}, 1), 0),
\end{equation}
where $x_{ij}$ denotes the $i, j-th$ pixel value, $m_{\bm{x}}$ denotes the $1^{st}$ percentile brightness value, and $M_{\bm{\tilde{x}}}$ denotes the $99^{th}$ percentile brightness of $\bm{x} - m_{\bm{x}}$. The resulting cropped and scaled images were then manually assigned a
distinct label from the categories \emph{Arc} (774 images), \emph{Discrete} (1102 images), \emph{Diffuse} (1400 images), \emph{Cloudy} (817 images), \emph{Moon} (585 images), \emph{Clear/No aurora} ((1082 images). These categories are described in~\cite{clausen2018automatic} as follows:

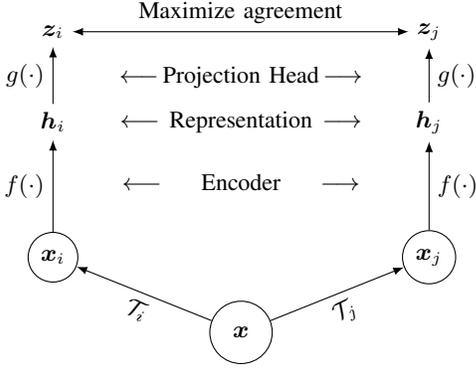
\begin{figure}[t]
\small
    \centering
\begin{tikzpicture}
    \node at (0, 2.4) (h)
    {$\longleftarrow\,$Projection Head$\,\longrightarrow$};
    \node at (0,1.8) (h) {$\longleftarrow\,$\hspace{0.25em}Representation\hspace{0.25em}$\,\longrightarrow$};
    \node at (0,1) (h)
    {$\longleftarrow\,$\hspace{1.5em}Encoder\hspace{1.5em}$\,\longrightarrow$};
    \node[draw, circle] at (0,-1) (x) {$\,~\bm{x}~\,$};
    \node[draw, circle] at (-2.5,0) (x1) {$\bm{x}_i$};
    \node[draw, circle] at (2.5,0) (x2) {$\bm{x}_j$};
    \node at (-2.5,1.8) (h) {$\bm h_i$};
    \node at (2.5,1.8) (c) {$\bm h_j$};
    \node at (-2.5,3) (hh) {$\bm z_i$};
    \node at (2.5,3) (cc) {$\bm z_j$};
    \path[->] 
        (x)  edge [>=latex] node[below,rotate=-25] {$\mathcal{T}_i$} (x1)
        (x)  edge [>=latex] node[below,rotate=25] {$\mathcal{T}_j$} (x2)
        (x1)  edge [>=latex] node[left,rotate=0] {$f(\cdot)$} (h)
        (x2)  edge [>=latex] node[right,rotate=0] {$f(\cdot)$} (c)
        (h)  edge [>=latex] node[left,rotate=0] {$g(\cdot)$} (hh)
        (c)  edge [>=latex] node[right,rotate=0] {$g(\cdot)$} (cc);
    \path[<->]
        (hh)  edge [>=latex] node[above,rotate=0] {Maximize agreement} (cc);
    \end{tikzpicture}
    \caption{A simple framework for contrastive learning of visual representations. Given a batch of images, for each sample $\bm x$, two distinct transformations are randomly applied ($\mathcal{T}_i$ and $\mathcal{T}_j$) to obtain two correlated views of $\bm x$, a `positive pair'.  A base encoder network $f(\cdot)$ and a projection head $g(\cdot)$ are then trained to identify the positive pairs among pairs constucted from all transformed samples in a minibatch. After training is completed, the encoder $f(\cdot)$ is applied to each $\bm x$ to obtain a representation $\bm h$ which can then be used for downstream tasks. Diagram adapted from \cite{chen2020simple}.}
    \label{fig:framework}
\end{figure}

\begin{itemize}
    \item \emph{Arc}: Showing one or multiple bands of aurora that stretch across the field-of-view; typically, the arcs have well-defined, sharp edges.
    \item \emph{Diffuse}: Large patches of aurora, typically with fuzzy edges.
    \item \emph{Discrete}: Auroral forms with well-defined, sharp edges, that are, however, not arc--like.
    \item \emph{Cloudy}: Dominated by clouds or the dome of the imager is covered with snow.
    \item \emph{Moon}: Dominated by light from the Moon.
    \item \emph{Clear/No aurora}: Images which show a clear sky (stars and planets are clearly visible) without the appearance of aurora.
\end{itemize}

Examples from each class are displayed in Figure \ref{fig:image-samples}.

\subsection{SimCLR}\label{section:simclr}

The Simple framework for Contrastive Learning of Representations (SimCLR) algorithm was introduced by Chen \emph{et. al.} in \cite{chen2020simple}. As the name suggests, SimCLR is a relatively simple and yet highly effective algorithm for unsupervised representation learning. The SimCLR framework consists of four components, described below and diagrammed in Figure \ref{fig:framework}:

\begin{itemize}
        \item A stochastic \emph{data augmentation module}. Given a sample $\bm x$, the data augmentation module randomly applies two transformations $\mathcal{T}_i, \mathcal{T}_j$ to $\bm x$, resulting in two distinct views $\bm{x}_i, \bm{x}_j$ of $\bm x$ that we consider a \emph{positive pair} in the context of the contrastive loss function described below. 
        \item A \emph{base encoder}. The base encoder is used to extract an initial representation of a transformed input $\mathcal{T}_k(\bm{x})$. In principle, this could be any machine learning model; in most applications a residual neural network (ResNet)~\cite{he2016deep}, sometimes pretrained on ImageNet, is used. 
        \item A \emph{projection head}. The projection head maps the output of the base encoder to the space where the contrastive loss function is to be applied. Again, in principle this could be any machine learning model; in practice the projection head is usually a shallow (one--layer) dense neural network. 
        \item A \emph{contrastive loss function} designed for the following contrastive prediction task: given a set of pairs of examples that includes a positive pair, identify the positive pair. 
\end{itemize}

The contrastive loss function used here is the \emph{normalized temperature--scaled cross--entropy loss} \cite{sohn2016improved,wu2018unsupervised,oord2018representation}. Letting $\csim(\bm{x}, \bm{y}) = \bm{x}^T\bm{y} /\norm{\bm{x}}||\bm{y}||$ (\emph{cosine similarity}), the loss for a positive pair of examples $\bm{x}_i$, $\bm{x}_j$ is given by

\begin{equation}\label{eq:nxent}
        \ell(\bm{x}_i, \bm{x}_j) = -\log\frac{\exp(\csim(\bm{z}_i, \bm{z}_j)/\tau)}{\sum_{k=1}^{2N}\mathbbm{1}_{[k\neq i]}\exp(\csim(\bm{z}_k, \bm{z}_i)/\tau)}.
\end{equation}

In Equation \ref{eq:nxent}, $\bm{z}_i, \bm{z}_j$ are the outputs of the projection head for two views $\bm{x}_i, \bm{x}_j$ of an input $\bm{x}$. $\tau$ is a temperature parameter, $\mathbbm{1}_{[k\neq i]}$ is an indicator function that evaluates to 1 if $i \neq k$, and the $2N$ examples represented in the sum in the denominator are obtained by using the $2(N-1)$ augmented pairs in a minibatch as negative examples. This mitigates any need to sample negative pairs
explicitly. The model is then trained to minimize

\begin{equation}
        \frac{1}{2N}\sum_{k=1}^N[\ell(2k-1, 2k) + \ell(2k, 2k-1)].
\end{equation}

The choice of transformations sampled from is believed to be a critical component in learning successful representations using the SimCLR algorithm. We opt for simplicity, and use only two: a random resized cropping of the input image and a random flip over a horizontal axis. The latter choice is motivated by the view that the location of auroral forms in the image will be relevant to the ultimate goal of obtaining a better understanding of the magnetospheric dynamics involved; a vertical flip causes the model to judge as similar forms that occur on
the left side of the image and forms that occur on the right, but a horizontal flip decorrelates the image pair while preserving the geographic location of the form. 

\begin{table*}[!t]
\begin{center}
\begin{tabular}{|c|c|c|c|c|c|c|}
\hline
Model & Params. & Top--1 Accuracy (\%)& Std. & Precision & Recall & F1--Score \\ \hline
Ridge + Inception (pretrained)\cite{clausen2018automatic} & 43m & 81.7 & 0.1 & 83.0$^{*}$ & 84.3$^*$ & 83.6$^*$  \\ \hline
ResNet18 (pretrained, supervised) & 12m & 84.7 & 0.7 & 85.6 & 85.2 & 85.3 \\ \hline 
SimCLR & 12m & $\bm{90.9}$ & $0.7$ & $\bm{91.6}$ & $\bm{91.5}$ & $\bm{91.5}$ \\ \hline
\end{tabular}

\caption{\textbf{Auroral image classification results.} Data in row 1 is taken from \cite{clausen2018automatic}. The starred entries in row 1 were not reported in the cited paper, and instead were calculated from a single fold confusion matrix provided in the paper. Supervised ResNet18 results included for comparison. Reported results are averages from $5$--fold cross--validation. To optimize comparability, the same folds are used for cross--validation as in \cite{clausen2018automatic}.}
\label{table:results}
\end{center}
\end{table*}

In \cite{chen2020simple}, it is noted that random color distortion, when paired with random cropping, is believed to be among the most important transformations to include in order to learn high--quality representations for image classification tasks. Our experience with this was mixed:  incorporating random color distortions did improve classification performance somewhat beyond the results documented here, but it also degraded the quality of the representations learned for clustering and other downstream tasks. We therefore omitted random color distortions along with other commonly included transformations such as random rotations and perspective shifts.

In all of our experiments, we use a projection head consisting of two fully connected linear layers with an intermediate ReLU activation and without biases; that is, $g(\bm{h}) = W_2(\max(0, W_1(\bm{h}))$, where $W_1$ and $W_2$ denote weight matrices. As an encoder, we use a lightweight ResNet18 model that is randomly initialized. The temperature hyperparameter $\tau$ of the loss function (c.f. Eq. \ref{eq:nxent}) was set to 0.5.

\subsection{Training}\label{section:training}

We train the SimCLR model for 100 epochs, using the Adam optimizer~\cite{kingma2014adam} with a fixed learning rate of 0.0003 and no weight decay. We use a minibatch size of 128. All models were implemented using the open source machine learning library PyTorch \cite{paszke2019pytorch} and trained using a single NVIDIA Titan Xp GPU.

\subsection{Classification}\label{section:classification}

After training, we fit an $\ell2$--regularized logistic regression classifier to the obtained representations. We use 5--fold cross--validation to obtain average classification performance as measured by a range of metrics; c.f. Table \ref{table:results}. For comparison, we also include results from fine--tuning an ImageNet--pretrained ResNet18. In order to insure maximum comparability between models
we use the same folds for cross--validation as in \cite{clausen2018automatic}. Linear classifiers were fit using Scikit-Learn~\cite{scikit-learn}.

\begin{figure}[H]
    \centering
    \includegraphics[width=0.48\textwidth]{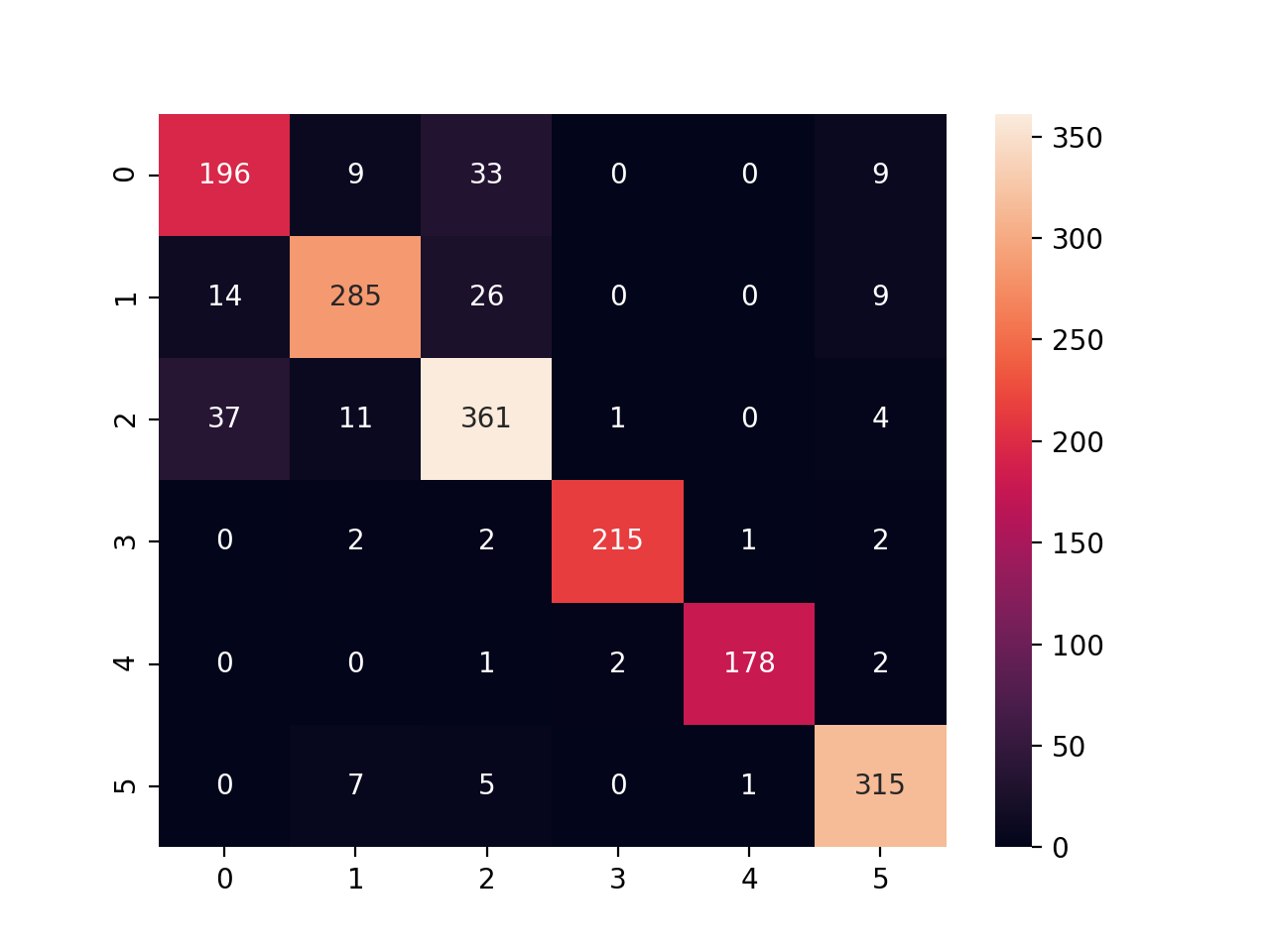}
    \caption{\textbf{Confusion matrix for a single fold.} Rows are ground--truth labels, columns are predicted labels.}
    \label{fig:clausen-conf}
\end{figure}

A confusion matrix for one of the held--out validation sets is presented in Figure~\ref{fig:clausen-conf}. Confusion matrices for the other folds are very similar.

\begin{figure}[htbp]
        \centering
        \begin{subfigure}[b]{0.15\textwidth}
                \centering
                \includegraphics[width=\textwidth]{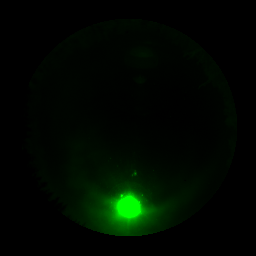}
                \caption{G: 4 S: 0}
        \end{subfigure}
        \hfill
        \begin{subfigure}[b]{0.15\textwidth}
                \centering
                \includegraphics[width=\textwidth]{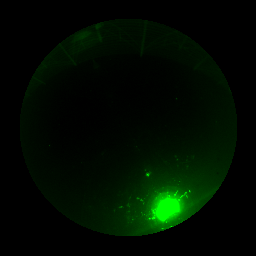}
                \caption{G: 4 S: 0}
        \end{subfigure}
        \hfill
        \begin{subfigure}[b]{0.15\textwidth}
                \centering
                \includegraphics[width=\textwidth]{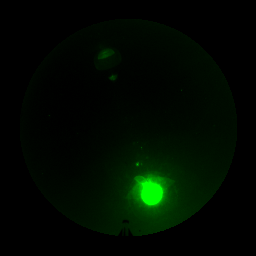}
                \caption{G: 4 S: 0}
        \end{subfigure}

        \begin{subfigure}[b]{0.15\textwidth}
                \centering
                \includegraphics[width=\textwidth]{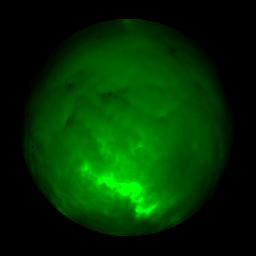}
                \caption{G: 3 S: 3}
        \end{subfigure}
        \hfill
        \begin{subfigure}[b]{0.15\textwidth}
                \centering
                \includegraphics[width=\textwidth]{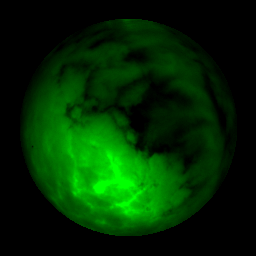}
                \caption{G: 3 S: 3}
        \end{subfigure}
        \hfill
        \begin{subfigure}[b]{0.15\textwidth}
                \centering
                \includegraphics[width=\textwidth]{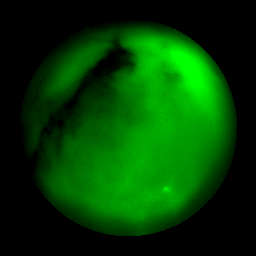}
                \caption{G: 3 S: 3}
        \end{subfigure}

        \begin{subfigure}[b]{0.15\textwidth}
                \centering
                \includegraphics[width=\textwidth]{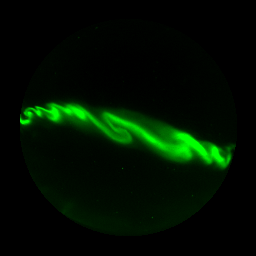}
                \caption{G: 2 S: 2}
        \end{subfigure}
        \hfill
        \begin{subfigure}[b]{0.15\textwidth}
                \centering
                \includegraphics[width=\textwidth]{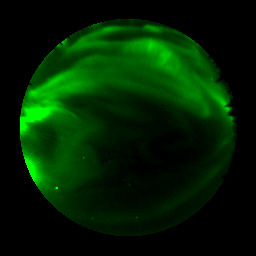}
                \caption{G: 2 S: 7}
        \end{subfigure}
        \hfill
        \begin{subfigure}[b]{0.15\textwidth}
                \centering
                \includegraphics[width=\textwidth]{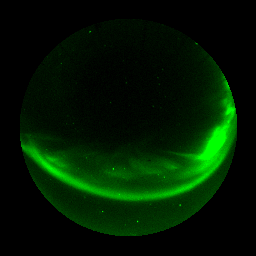}
                \caption{G: 2 S: 11}
        \end{subfigure}
        \caption{A random sample of images from three ground--truth categories and their cluster labels. `G' indicates the ground--truth label (0='Arc aurora', 1='Diffuse aurora', 2='Discrete aurora', 3='Cloudy', 4='Moon', 5='Clear/No aurora'); `S' the category assigned by the K-means algorithm to the learned representation of the image. In rows 1 and 2, the categories largely agree. In row 3, the `Discrete aurora' category, which can be seen in the sample to contain very different auroral forms, is split over several distinct clusters.}\label{figure:clausen-kc-comparison} 
\end{figure}

\subsection{Clustering}\label{section:clustering}

\begin{figure}[t]
        \centering
        \begin{subfigure}[b]{0.15\textwidth}
                \centering
                \includegraphics[width=\textwidth]{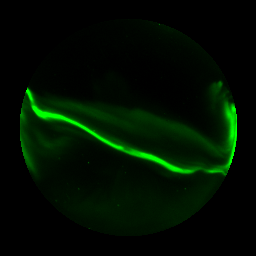}
                \caption{G: 2 S: 2}
        \end{subfigure}
        \hfill
        \begin{subfigure}[b]{0.15\textwidth}
                \centering
                \includegraphics[width=\textwidth]{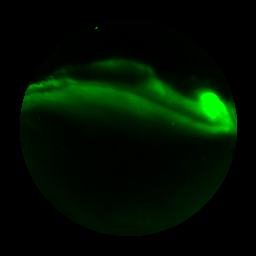}
                \caption{G: 2 S: 2}
        \end{subfigure}
        \hfill
        \begin{subfigure}[b]{0.15\textwidth}
                \centering
                \includegraphics[width=\textwidth]{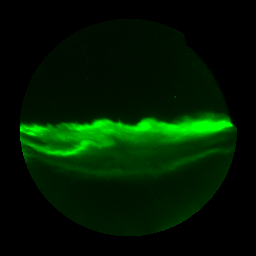}
                \caption{G: 2 S: 2}
        \end{subfigure}

        \begin{subfigure}[b]{0.15\textwidth}
                \centering
                \includegraphics[width=\textwidth]{images/visualization-sample/clausen/02446.png}
                \caption{G: 2 S: 7}
        \end{subfigure}
        \hfill
        \begin{subfigure}[b]{0.15\textwidth}
                \centering
                \includegraphics[width=\textwidth]{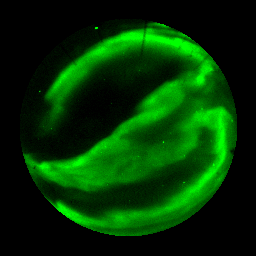}
                \caption{G: 2 S: 7}
        \end{subfigure}
        \hfill
        \begin{subfigure}[b]{0.15\textwidth}
                \centering
                \includegraphics[width=\textwidth]{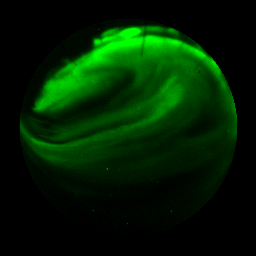}
                \caption{G: 2 S: 7}
        \end{subfigure}

        \begin{subfigure}[b]{0.15\textwidth}
                \centering
                \includegraphics[width=\textwidth]{images/visualization-sample/clausen/03545.png}
                \caption{G: 2 S: 11}
        \end{subfigure}
        \hfill
        \begin{subfigure}[b]{0.15\textwidth}
                \centering
                \includegraphics[width=\textwidth]{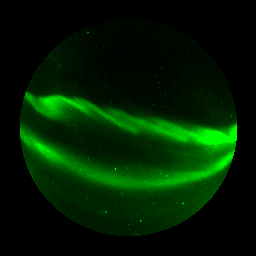}
                \caption{G: 2 S: 11}
        \end{subfigure}
        \hfill
        \begin{subfigure}[b]{0.15\textwidth}
                \centering
                \includegraphics[width=\textwidth]{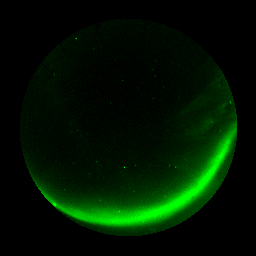}
                \caption{G: 2 S: 11}
        \end{subfigure}
        \caption{A random sample of images from the `Discrete aurora' category in specific cluster assignments. `G' indicates the ground--truth label; `S' the cluster label (c.f. Figure~\ref{figure:clausen-kc-comparison}). The images in each row share characteristics that distinguish them from the images in the other rows.}\label{fig:clausen-kc-comparison-2}  
\end{figure}

After training, the $K$--means clustering algorithm was applied to cluster the learned representations. By comparing average silhouette scores for $k \in \{3,\dots,15\}$, we find evidence that the number of clusters present in the data is greater than the number of ground--truth labels currently used to classify the dataset. Specifically; we attain a maximum average silhouette score of $0.212$ for $k=12$. For comparison, using the ground--truth labels as cluster assignments yields an average silhouette score of $0.011$. Significantly, while some clusters align closely with the ground--truth classifications (the classes \emph{Cloudy} and \emph{Moon}, for example), other less well--defined ground--truth
classes (in particular, the Discrete class) are split among various clusters; c.f. Figure~\ref{figure:clausen-kc-comparison}, \ref{fig:clausen-kc-comparison-2}. Moreover, random samples drawn from the learned clusters exhibit clear qualitative similarities; c.f. Figure~\ref{fig:clausen-kc-comparison-2}. 

\section{Discussion}\label{section:discussion}

There are several distinctions to be made between the approach outlined here and the standard SimCLR model. As mentioned previously, we greatly reduce the number of transformations sampled from and nevertheless obtain high performance. We train on a small dataset using a lightweight model with 12m parameters, and improve on the current
state--of--the--art, suggesting that while the conventional wisdom is that to learn effective representations unsupervised models should be more highly parameterized than their supervised counterparts, this may in fact not generally be the case.

In \cite{syrjasuo2011numeric}, it is noted that classification error rates of less than 10\% are likely to be sufficient for most practical purposes; in particular, for automatic labeling of auroral images to take place on a large scale, thereby enabling statistical studies of auroral images not previously possible. The results presented in Section \ref{section:classification}, Table~\ref{table:results}, and Figure \ref{fig:clausen-conf} demonstrate that the approach described here meets this criteria. Moreover, the model is relatively lightweight, using slightly less than 12m parameters
and requiring less than 110MB of space to train to convergence. We obtain our state--of--the--art results in spite the fact that our model does not rely on ImageNet pretraining, the usual approach when working with similarly sized datasets, thus removing another potential source of bias\cite{geirhos2018imagenet}.

The results presented in Section \ref{section:clustering} align with the general fact that the morphology of auroral forms is not well understood. The approach outlined here leads to representations that provide a much finer categorization of auroral forms, with clear qualitative distinctions between categories. This suggests that unsupervised learning may well have an important role to play in improving our understanding of the connection between what is viewed from the ground and the dynamics and coupling of Earth’s magnetosphere and upper atmosphere. 

\section{Conclusions and Future Work}\label{section:conclusion}

In this paper, we have presented a novel approach based on adapting the SimCLR model introduced in \cite{chen2020simple} to the practical challenge of auroral image classification. Our approach leads to state--of--the--art results as measured by a range of classification metrics, surpassing an important threshold for practical utility while requiring less than 25\% of the parameters of recent benchmark models and without relying on ImageNet
pretraining. Our approach provides preliminary evidence to suggest that current choices of ground--truth labels for auroral images are likely overly coarse. We also demonstrate that the guidelines for learning effective representations put forth in \cite{chen2020simple}
may be ideal for conventional object recognition tasks, but may require adjustment for data that does not share the same general characteristics. More research is needed to better understand the connection between the representations
learned by our models and the magnetospheric processes involved. This will be a focus of future work.

%
%

\bibliographystyle{IEEEtran}
\bibliography{icmla2021}
\end{document}